\documentclass[conference, letterpaper]{IEEEtran}
\ifCLASSINFOpdf
\else
\fi
\usepackage{subcaption}

\ifCLASSINFOpdf
   \usepackage[pdftex]{graphicx}
\else
\fi

\usepackage[cmex10]{amsmath}
\usepackage{color}
\usepackage{fancyhdr}

\renewcommand{\thispagestyle}[2]{}

\fancypagestyle{plain}{
        \fancyhead{}
        \fancyhead[C]{first page center header}
        \fancyfoot{}
        \fancyfoot[C]{first page center footer}
}
\usepackage{graphicx}
\usepackage{amsmath}
\usepackage{relsize}
\usepackage{algorithm}
\usepackage{algorithmicx}
\usepackage{program}
\usepackage[noend]{algpseudocode}
\usepackage[colorlinks=true,
            linkcolor=red,
            urlcolor=blue,
            citecolor=blue]{hyperref}
\begin{document}

\title{FSL-BM: \underline{F}uzzy \underline{S}upervised \underline{L}earning with \underline{B}inary \underline{M}eta-Feature for Classification\\}
\author{
  \IEEEauthorblockN{
   Kamran Kowsari\IEEEauthorrefmark{1}, 
    Nima Bari\IEEEauthorrefmark{2}, 
    Roman Vichr\IEEEauthorrefmark{3} and 
    Farhad A. Goodarzi\IEEEauthorrefmark{4}}
  \IEEEauthorblockA{
    \IEEEauthorrefmark{1}Department of Computer Science, University of Virginia,
Charlottesville, VA USA\\
      Email: kk7nc@virginia.edu}
    \IEEEauthorblockA{\IEEEauthorrefmark{2}Department of Computer Science,
The George Washington University\\
      Email: nbari@gwu.edu}
          \IEEEauthorblockA{\IEEEauthorrefmark{3}Data Mining \& Surveillance \& Metaknowledge Discovery
\\
Email: namorvi@gmail.com}
\IEEEauthorblockA{\IEEEauthorrefmark{4}Department of Mechanical \& Aerospace Engineering,
The George Washington University,\\
      Email: fgoodarzi@gwu.edu}}
\maketitle

\begin{abstract}
This paper introduces a novel real-time Fuzzy Supervised Learning with Binary Meta-Feature~(FSL-BM) for big data classification task. The study of real-time algorithms addresses several major concerns, which are namely: accuracy, memory consumption, and ability to stretch assumptions and time complexity. Attaining a fast computational model providing fuzzy logic and supervised learning is one of the main challenges in the machine learning. In this research paper, we present FSL-BM algorithm as an efficient  solution of supervised learning with fuzzy logic processing using binary meta-feature representation using Hamming Distance and Hash function to relax assumptions. While many studies focused on reducing time complexity and increasing accuracy during the last decade, the novel contribution of this proposed solution comes through integration of Hamming Distance, Hash function, binary meta-features, binary classification to provide real time supervised method. Hash Tables~(HT) component gives a fast access to existing indices; and therefore, the generation of new indices in a constant time complexity, which supersedes existing fuzzy supervised algorithms with better or comparable results. To summarize, the main  contribution of this technique for real-time Fuzzy Supervised Learning is to represent hypothesis through binary input as meta-feature space and creating the Fuzzy Supervised Hash table to train and validate model.\\
\end{abstract}

\begin{IEEEkeywords} 
Fuzzy Logic; Supervised Learning; Binary Feature; Learning Algorithms; Big Data; Classification Task
\end{IEEEkeywords}


%
\IEEEpeerreviewmaketitle

\section{Introduction and Related Works}

Big Data Analytics has become feasible as well as recent powerful hardware, software, and algorithms developments; however, these algorithms still need to be fast and reliable~\cite{brazdil2008metalearning}. The real-time processing, stretching assumptions and accuracy till remain key challenges. Big Data Fuzzy Supervised Learning has been the main focus of latest research efforts~\cite{fatehi2017application}. Many algorithms have been developed in the supervised learning domain such as Support Vector Machine (SVM) and Neural Networks. Deep Learning techniques such as Convolutional Neural Networks~(CNN),  Recurrent Neural Networks~(RNN),  Deep Neural Networks~(DNN), and Neural Networks~(NN) are inefficient for fuzzy classification tasks in binary feature space\cite{qiu2017empirical,hinton2006reducing}, but Deep learning could be very efficient for multi-class classification task~\cite{kowsari2017HDLTex}. In fuzzy Deep neural networks, the last layer of networks~(output layer) is activated by Boolean output such as sigmoid function. Their limitation was demonstrated in their inability to produce reliable results for all possible outcomes. Time complexity, memory consumption, the accuracy of learning algorithms and  feature selection remained as four critical challenges in classifier algorithms.\\
The key contribution of this study  is providing a solution that addresses all four critical factors in a single robust and reliable algorithm while retaining linear processing time.

Computer science history in the field of machine learning has been shown significant development particularly in the area of Supervised Learning~(SL) applications~\cite{ashfaq2017fuzziness}. Many supervised learning applications and semi-supervised learning algorithms were developed with Boolean logic rather than using Fuzzy logic; and therefore, these existing methods cannot cover all possible variations of results. Our approach offers an effective Fuzzy Supervised Learning (FSL) algorithm with a linear time complexity. Some researchers have attempted to contribute in their approach to Fuzzy Clustering and utilizing more supervised methods than unsupervised. Work done in 2006 and in 2017,~\cite{jiang2006fuzzy,chen2017new} provided new algorithm with Fuzzy logic implemented in Support Vector Machine (SVM), which introduced a new fuzzy membership function for nonlinear classification. In the last two decades, many research groups focused on Neural Networks using Fuzzy logic~\cite{chen2014fuzzy} or neuro-fuzzy systems~\cite{sajja2017computer}, and they used several hide layer and  .  In 1992, Lin and his group worked on the Fuzzy Neural Network (FNN). However, their contribution is besed on outlined in the back-propagation algorithm and real time learning structure~\cite{lin1992real}. Our work focuses on approach of mathematical modeling of binary learning with hamming distance applied to supervised learning.

Between 1979 and 1981, NASA\footnote{The National Aeronautics and Space Administration}  developed Binary Golay Code (BGC) as an error correction technique by using the hamming distance~\cite{thompson1983error,west2008commercializing}. The 1969 goal of these research projects was an error correction using Golay Code for communication between the  International Space Station and Earth. Computer  scientists and electrical engineers used fuzzy logic techniques for Gilbert burst-error-correction over radio communication~\cite{bahl1969gilbert,yu2012golay}. BGC utilizes 24 bits, however, a perfected version of the Golay Code algorithm works in a linear time complexity using 23 bits~\cite{rangare2016review},~\cite{berkovich2007method}. The algorithm used and implemented in this research study was inspired by the Golay Code clustering hash table~\cite{kowsari1construction, bari2015novel, berkovich2007method,kowsari2014investigation}. This research offers two main differences and improvements: i) it works with $n$ features whereas Golay code has a limitation of 23 bits ii) our method utilizes supervised learning while Golay Code is an unsupervised algorithm which basically is a Fuzzy Clustering method. The Golay code generate hash table with  six indices for labelling Binary Features (BF) as fuzziness labeled but FSL-BM is supervised learning is induced techniques of encoding and decoding into two labels  or sometimes fuzzy logics classifiers by using probability or similarity. Between 2014 and 2015, the several studies addressed on using the Golay Code Transformation Hash table~(GCTHT) in constructing a 23-bit meta-knowledge template for Big Data Discovery which allows for meta-feature extraction for clustering Structured and Unstructured Data (text-based and multimedia)~\cite{bari201423,bari2015novel}. In 2015, according to~\cite{kowsari1construction}, FuzzyFind Dictionary (FFD), is generated by using GCTHT and FuzzyFind dictionary is improved from \%~86.47~(GCTHT) to \%98.2~percent~\cite{kowsari1construction} .In this research our meta-features and feature, selection are similar to our previous work, which is done by Golay Code Clustering, but now we introduce a new algorithm for more than 23 features. Furthermore, existing supervised learning algorithms are being challenged to provide proper and accurate labeling~\cite{kamishima2003clustering} for unstructured data.\\ Nowadays, most large volume  data-sets are available for researchers and developers contain data points belonging to more than a single label or target value.  Due to the limited time complexity and memory consumption, existing fuzzy clustering algorithm such as genetic fuzzy learning~\cite{russo2000genetic} and fuzzy C-means~\cite{bezdek1984fcm} aren't very applicable for Big Data. 
Therefore, a new method of fuzzy supervised learning is needed to process, cluster, and assign labels to unlabeled data using a faster time complexity, less memory consumption and more accuracy for unstructured datasets. In short, new contributions and the unique features of the algorithms proposed in this paper are an efficient technique of Fuzziness learning, linear time complexity, and finally powerful prediction due to  robustness and complexity. The baseline of this paper is as follows: Fuzzy Support Vector Machine(FSVM)~\cite{qin2017nested} and Original  Support Vector Machine(SVM).\\

This paper is organized with the following topics respectively: section~\ref{FuzzyLogic}:~Fuzzy Logic for Machine Learning, section~\ref{PreProcessing}:~Pre-Processing including section~\ref{Meta}:~Meta-Knowledge. section~\ref{MetaL}:~Meta-Feature Selection,  section~\ref{Supervised}:~Supervised 
Learning including section~\ref{PSupervised}:~Pipeline of Supervised Learning by Hamming Distance and how we train our model, and finally, section~\ref{eval})~evaluation of model; and finally, section~\ref{Results}:~experimental results.
\begin{figure}[b] \label{fig:fuzzy}
  \centering
    \includegraphics[width=\columnwidth]{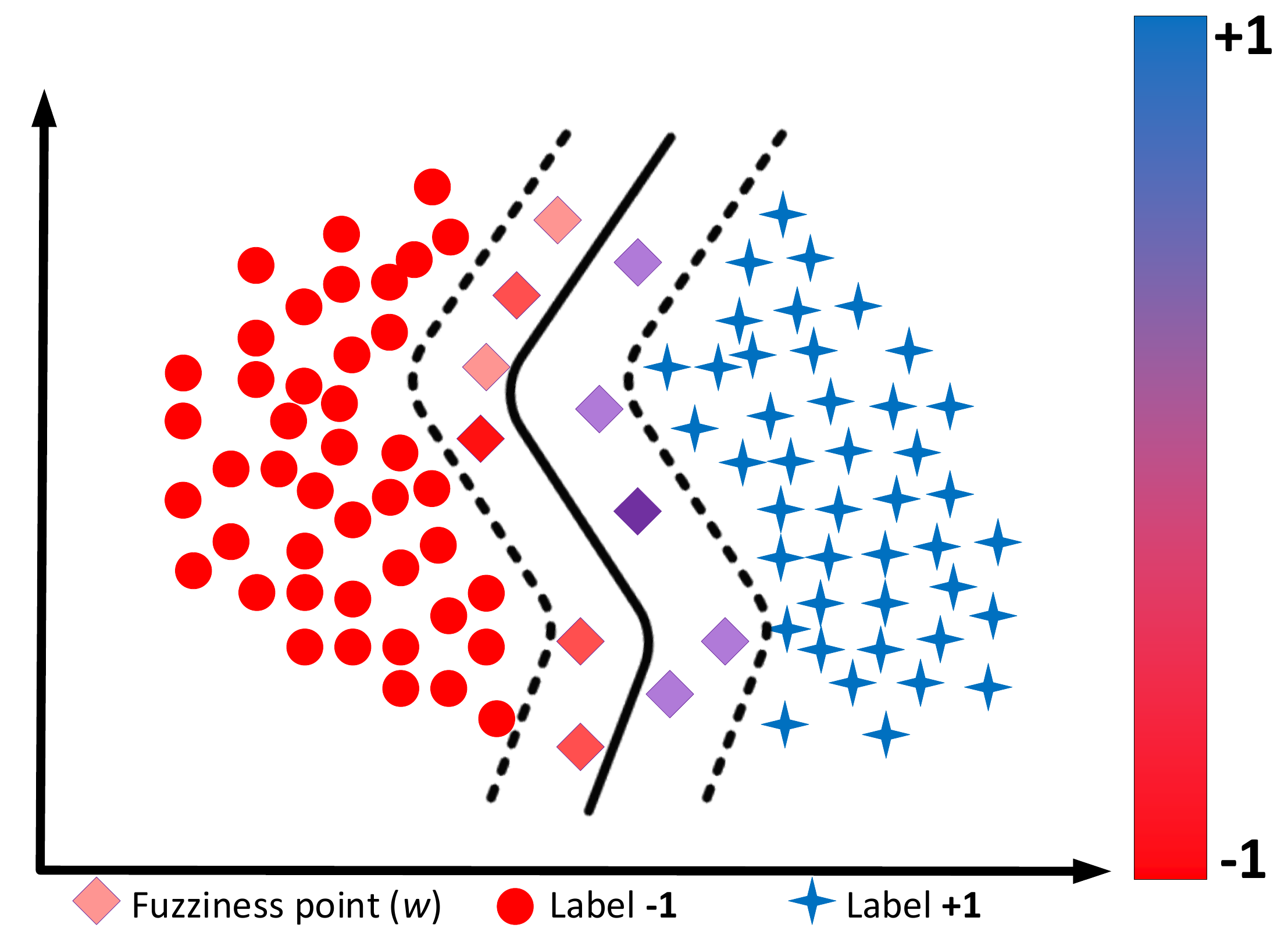}
  \caption{How Fuzzy Supervised Learning works on fuzzy datasets. In this figure $w$ indicates the percentage of fuzziness with means if $w=[0.2,0.8]$ that data point belongs $20$ percents to label $-1$ and $80\%$ belongs to label $+1$ . }\label{fig:fuzzy}
\end{figure}

\begin{figure*}
  \centering \includegraphics[width=\textwidth]{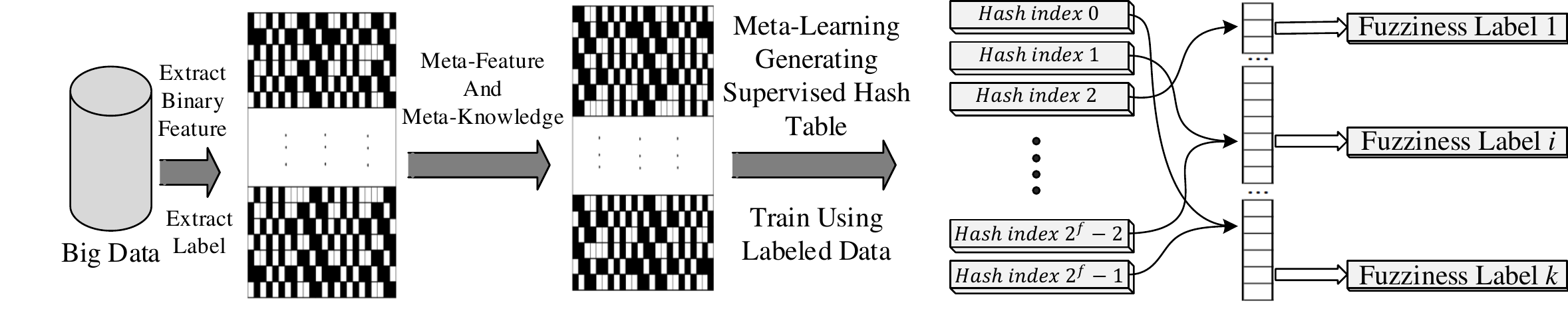}
\caption{This figure indicates how generating FSL-BM. From left to right, Extraction of Binary Input from unstructured big data; and then, we generate Meta-Feature or meta-knowledge; and finally, Fuzzy Hash table is created to use in   Supervised Learning }\label{pip}
\end{figure*}

\section{Fuzzy Logic for Machine Learning}\label{FuzzyLogic}
Fuzzy logic methods in machine learning are more popular among  researchers~\cite{wieland2017combining,prabu2016fuzzy} in comparison to Boolean and traditional methods. The main difference between the Fuzziness method in clustering and classification for both fields of supervised and unsupervised learning is that each data point can be belong to more than one cluster. Fuzzy logic, in our case, is extended to handle the concept of partial truth, where the truth-value may range between completely true [1] and false [0]. We make the claim that such an approach is suited for the proposed binary stream of data meta-knowledge representation~\cite{gama2010knowledge,gama2007learning}, which leads to meta-features. Therefore we apply Fuzzy logic as a comparative notion of truth (or finding the truth) without the need to represent fully the syntax, semantics, axiomatization, truth-preserving deduction, and still reaching a degree of completeness~\cite{hohle2012non}. We extend the many-valued logic~\cite{zalta2003stanford,forrest1996identity,logic2006stanford,pinto2014framework} based on the paradigm of inference under vagueness where the truth-value may range between completely true (correct outcome, correct label assignment) and false (false outcome, opposite label assignment), and at the same time the proposed method handles partial truth, where the label assignment can be either $\{1, 0\}$.

Through an optimization process of discovering meta-knowledge and determining of meta-features, we offer binary output representation as input into a supervised machine learning algorithm process that is capable of scaling. Each unique data point is assigned to a binary representation of meta-feature which is converted consequently into hash keys that uniquely represent the meta-feature presented in the record. In the next step, the applied hash function selects and looks at the supervised hash table to assign an outcome, which is represented by assigning the correct label. The fuzziness is introduced through hash function selection of multiple (fuzzy) hash representations~\cite{forrest1996identity,logic2006stanford,pinto2014framework}.

The necessary fuzziness compensates for the inaccuracy in determining the meta-feature and its representation in the binary data stream. As we represent these meta-features as binary choice of $\{1, 0\}$, we provide binary output of classification outcome as $\{1, 0\}$ through the designation of labels \cite{forrest1996identity,logic2006stanford,pinto2014framework}. There must be some number of meta-features ($n = i$) such that a record with n meta-features counts  with "result m" whilst a record with $n+1$ or $n-1$ does not. Therefore, there must be some point where the defined and predicted output (outcome) ceases. Let $\exists~n (\dots n \dots)$ assert that some number n satisfies the condition …n…. Therefore, we can represent the sequence of reasoning as follows,
\begin{equation}\label{eq1}
F_{a_1} \sim ~\forall~~ n, 
\end{equation}
\begin{equation}
F_{a_n} \exists (F_{a_n} \sim F_{a_{n+1}}) F_{a_i},               
\end{equation}

where $i$ can be arbitrarily large. If we paraphrase the above expressions with utilization of Hamming Distance~(HD), there must be a number of meta-features~$(n = i)$ such that a record with $n$ meta-features counts with result $m$ while a records with~$(n+ HD)$ or~$(n-HD)$ does not exist. Whether the argument is taken to proceed by addition or subtraction\cite{cargile1969sorites,Malinowski200713}, it completely depends on how one views the series of meta-features \cite{dinis2017old}. This is the key foundation of our approach that provides background to apply and evaluate many valued truth logics with the standard two value logic (meta-logic), where the truth and false, i.e., yes and no, is represented within the channeled stream of data. The system optimizes (through supervised) training on selection of meta-features to assert fuzziness of logic into logical certainty; thus we are combining the optimization learning methods through statistical learning (meta-features) and logical (fuzzy) learning to provide the most efficient machine learning mechanism \cite{zalta2003stanford,forrest1996identity,logic2006stanford,pinto2014framework}.

The Figure~\ref{fig:fuzzy} indicates how fuzzy logics works on supervised learning for two classes. This figure indicates that red circle assigned only in label~$-1$ and blue stars belong to label~$+1$, but diamond shape there dose not have specific color which means their color is between blue and red. if we have $k$ number of classes or categories, the data points can be belonging to $k$ different categories.
\begin{equation}
    W = [w_1,w_2,...,w_k]
\end{equation}
\begin{equation}
    \sum_{i=0}^kw_i = 1
\end{equation}
Where $k$ is number of categories, $W$ is labels of data points and $w_i$ is percentages of labelling for class $i$. 
\section{Pre-Processing}\label{PreProcessing}
Regarding the hash function, the order of the features is critical for this learning techniques as a feature space~\cite{kowsari1construction,kowsari2014investigation,yammahi2014efficient}. Therefore, we use a process for feature selection that consists of meta-feature collection, meta-feature learning, and meta-feature selection. The $n$ feature that build the meta-knowledge template technique offers unique added value in that it provides clusters of interrelated data objects in a fast and linear time. The meta-knowledge template is a pre-processing technique built with each feature that can be assigned with either a yes or no as binary logics. In other words, given a template called  $F={f_1,f_2,f_3,...,f_n}$, $f$ is a single feature representing a bit along the $n$-bit string. It is good to indicate that developing meta-knowledge is associated with the quality methodology associated with ontology engineering. Ontology aggregates a common language for a specific domain while specifying definitions and relationships among terms. It is also important to indicate that the development of the meta-knowledge template is by no means done randomly. This opportunity seems to be unique and unprecedented. In the following sections, we will explain the process that constitutes the building of the meta-knowledge based on specific feature selections that defines the questions of meta-knowledge.

\subsection{Meta-knowledge}\label{Meta}
The definition of meta-knowledge is extracting knowledge from feature representation and also we can define it as perfect feature extraction and pre-selected knowledge from unstructured data~\cite{evans2011metaknowledge,handzic2004knowledge,Kazem}. The meta-knowledge or perfect feature extraction allows the deep study of feature for the purpose of more precise knowledge. Meta-knowledge can be utilized in any application or program to obtain more insightful results based on advanced analysis of data points.\\n the early 1960s, researchers were challenged to find a solution for a large domain specific knowledge~\cite{davis1984meta}. The goal to collect and utilize knowledge from these large data-repositories has been a major challenge, as a result meta-knowledge systems have been developed to overcome this issue. The problem of those to represent this knowledge remained a research question for researchers to develop. Therefore, our presented approach of meta-knowledge template with $n$-features can significantly provide easiness and speed to process large data sets.
\begin{algorithm}[h]
\caption{Generating List of Meta-features}\label{al1}
  \begin{algorithmic}[1]
\For{\texttt{i = 1 to $2^f-1$}}
\For{\texttt{j= 1 to $\sum_{i=0}^h \begin{pmatrix} f  \\ i  \end{pmatrix}$}}
        \State \#\# statistical meta-feature determination
        
\If{$(U_c,U_j) = \sum_{k=1}^f Err_k \leq \min(e)$}
\State ~ 
\Comment{Stattistical Error}
\If{$\eta_{i} ~is ~Null$}
\State $\eta_{i,j} \leftarrow \eta_{i,new}$ \Comment{Create meta-feature (domain knowledge)}
\Else
\State $\eta_{i,j} \leftarrow \eta_{i,j}+   \Psi$ \Comment{Add tested meta-feature}
        \EndIf
        \State \textbf{End of if}
        \EndIf
        \State \textbf{End of if}
      \EndFor
      \State \textbf{End of For}
      \EndFor
      \State \textbf{End of For}
  \end{algorithmic}
\end{algorithm}
\subsection{Meta-learning}\label{MetaL}
According to~\cite{vilalta2004using}, Meta Learning is a very effective technique for solving support data mining. In regression and classification problems  Meta Feature (MF) and Meta Learning algorithms have been used for application in the data mining and machine learning domain. It is very important to mention that the results obtained from data mining and machine learning are directly linked to the success of well-developed meta-learning model. In this research, we define Meta-learning as the process which helps the selected features use the right machine learning algorithms to build the meta-knowledge. The combination of machine learning algorithms and the study of pattern recognition allows us to study the meta-features correlations and the selections of the most result/goal-indicating features.
\begin{figure}[h]
  \centering
    \includegraphics[width=\columnwidth]{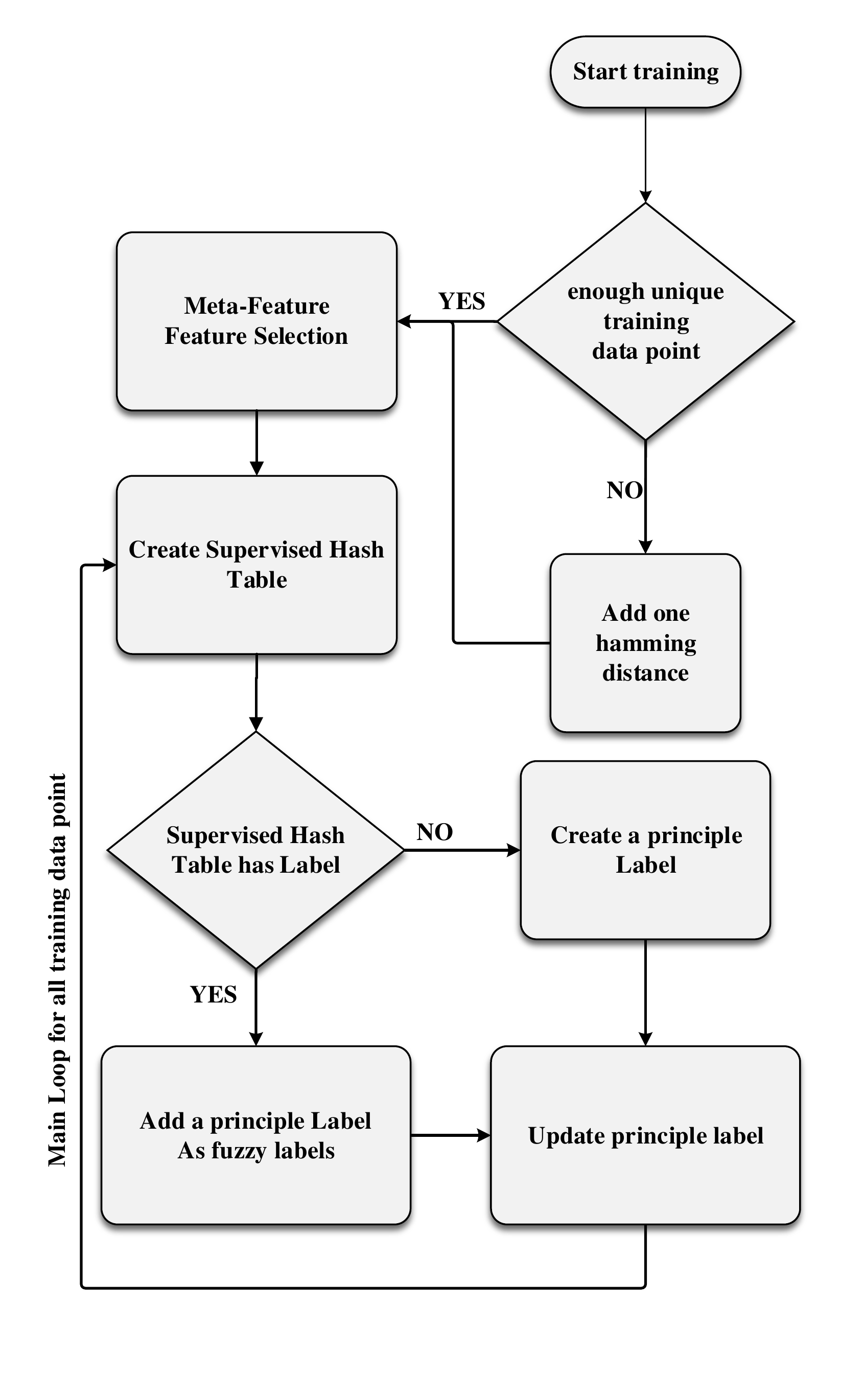}
  \caption{Pipeline of generating supervised training hash table using Hamming Distance }\label{workflowedge}
\end{figure}
\section{Supervised Learning}\label{Supervised}
There are generally three popular learning methods in the machine learning community: supervised learning, unsupervised learning, and semi-supervised learning. Unsupervised learning or data clustering by creating labels for unlabeled data points such as Golay Code, K-means, weighted unsupervised  and etc.~\cite{alassaf2015automatic,kowsari2016W,qazanfari2017efficient}. 

In Supervised Learning, more than 80 percent of the data points are used for training purposes, and the rest of data points will be used for testing purposes or evaluation of the algorithm such as Support Vector Machine (SVM), and Neural Network. Semi-supervised learning uses label generated by supervised learning on part of the data to be able to label the remaining data points~\cite{chapelle2009semi,chapelle2006continuation,chapelle2006branch}. The latter is a combination of supervised and unsupervised learning.
Overall, the contribution of this paper is shown in Fig.~\ref{pip} which is conclude to meta-feature learning in pre-processing step and the input feature is ready for learning algorithm in follows.

\begin{figure*}
  \centering \includegraphics[width=\textwidth]{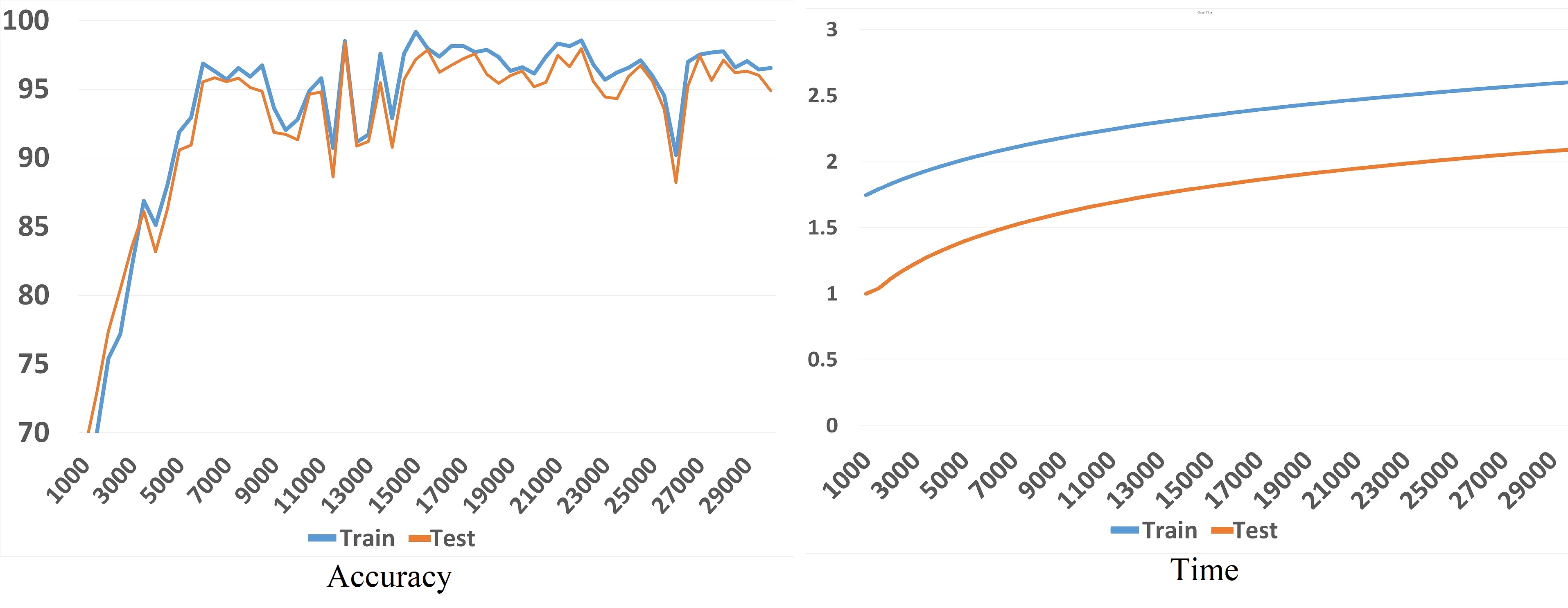}
  \caption{Left Figure: In this graph, Accuracy has been shown for online data stream which is increasing over by larger volume of data training. Left Figure) training time and testing time both is linear but generating FSL-BM is near to linear but validations test is faster due to use indexing of hash table \textbf{(Time is in log of second)} }\label{random}
\end{figure*} 
\subsection{Pipeline of Supervised Leaning using Hamming Distance}\label{PSupervised}
In the pipeline of this algorithm (Fig.~\ref{workflowedge}), all possible combinations of input binary features are created and the algorithm improves the training matrix by using hamming distance and ultimately improves the results by meta-feature selection and meta-knowledge discovery. As show in Figure 1, the algorithm is divided into two main parts; i) the training algorithm, which entails feature selection, hamming distance detection, and updates the training matrix, ii) testing the Hash function which is included in the meta-feature category; the critical feature order that converts the testing input to indices while each index has at least one or more label.

An explicit utilization for all available data points is not feasible. Supervised Hash Table(SHT) is a hash table with $2^f$  elements where $f$ is the number of binary feature~$f \in \{0 ... 2^f-1\}$ indices. The SHT elements are created by Hamming Distance of training data sets from zero to $2^f-1$. In equation \ref{eq2} , $h$ is the value of Hamming Distance which can be either $\{1,2,3\}$ or more depending on the number of training data points, and $f$ is number of features. The segmentation of the stream data sets can be {20$\dots$32} bits, and $\phi$ is the number of training data points.
\begin{align}\label{eq2}
\phi \mathlarger{‎‎\sum_{k=0}^h}{\begin{pmatrix} f  \\ k  \end{pmatrix}} = \phi\bigg( \begin{pmatrix} f  \\ 1  \end{pmatrix} + \begin{pmatrix} f  \\ 2  \end{pmatrix}+~\cdots~+ \begin{pmatrix} f  \\ h  \end{pmatrix}\bigg)
\end{align}

\subsection{Data Structure}
Data structure is one-implementation criteria for learning algorithms. According to ~\cite{kowsari2014investigation,kowsari1construction} the Golay code, Golay Code Transformation Matrix, Golay Code Clustering Hash Table, FuzzyFind Dictionary, and supervised Hash Table use the Hash table which can be the most efficient method for direct access to data by constant time complexity. Hash table is the most efficient techniques for knowledge discovery and it gives us constant time complexity to have easy and linear access to indices. On the other hand, Hash function can convert any unstructured or structured data input into binary features. This data structure is used in order to reduce computational time complexity in the supervised learning algorithm.
 \begin{algorithm}[]
\caption{Generating Supervised Hash Table}\label{al2}
  \begin{algorithmic}[1]
\For{\texttt{$c = 1$ to $\phi$}}
\For{\texttt{$j= 1$ to $\sum_{i=0}^e \begin{pmatrix} f  \\ i  \end{pmatrix}$}}
        \State \#\# statistical meta-feature determination
        
\If{$HD(U_c,U_j) = \sum_{k=1}^f \frac{\delta (C_{c k}-C_{j k})}{f}\leq e$}
\State ~ 
\Comment{HD is Hamming Distance}
\If{$w_i$ is $Null$  }
\State $w_{i,j} ~~\leftarrow w_{i,new} ~~ $
\Else
\State $w_{i,j^*} ~~\leftarrow w_{i,j} + \zeta ~~ $
\State ~ 
\Comment{Fuzzy logic oo training model}

        \EndIf
        \State \textbf{End of if}
        \EndIf
        \State \textbf{End of if}
      \EndFor
      \State \textbf{End of For}
      \EndFor
      \State \textbf{End of For}
  \end{algorithmic}
\end{algorithm}
\subsection{Hamming Distance}
Hamming Distance (HD) is used to measure the similarity of two binary variables~\cite{choi2010survey}. The Hamming Distance between two Code-word is equal to the number of bits in which they differ; for example: Hamming distance of~$0000\; 0100\; 0000\; 1000\; 1001\; (16521)$ and~$0000\; 0010\; 1110\; 0100\; 1111\; (15951)$ is equal~to~$9$. In the proposed algorithm, we use the values for HD of~$1$,~$2$,~and~$3$.This algorithm can Handel larger volume of data using fuzzy logics~(depends on the hardware , the algorithm is run on it). The number of bits is represented as binary meta-feature. In our algorithm, generate~$n$ bits as feature space(e.g. for~32 bits,~4-billon unique data points will be generated). In this paper, we test our algorithm with 24 binary input which means $2^24$ which has nearly 16 million unique records. 
\begin{align}
HD(U_c , U_j) = \sum_{k=1}^f \frac{\delta (C_{c~k} - C_{j~k})}{f}
\end{align}

\subsection{Generating the Supervised Hash Table}
Generating the Supervised Learning Hash Table is a main part of this technique. In this section, the main loop is given from 0 to all training data points for creating all possible results. The calculation of the Hamming Distance $e$ can be 2, 3 or even more for a large number of features and small portion of training data points. After calculating the hamming distance, the Supervised Hash Table is updated with labels.  Regarding Algorithm \ref{al2}, the main loop for all training data points, $\phi$ is as follows:
\begin{equation}\label{e1}
\phi(C_c,C_j)=\sum_{k=1}^f \frac{\delta (C_{c k}-C_{j k})}{f}\leq e
\end{equation}
\begin{equation}\label{ez}
w_{i,j^*} ~~\leftarrow w_{i,j} + \zeta
\end{equation}
Equation \ref{e1} is the hamming distance of features and $e$ indicates $max$ value of HD. If a label is assigned two or more labels, that vector of Supervised learning Hash Table keeps all of the labels in hash function, meaning the record uses fuzziness labeling.

\section{Evaluating Model}\label{eval}
In Supervised Learning using Hamming Distance techniques, the Supervised hash table during the training part. This Hash table contains all possible feature's input if enough data used for training. For evaluating the trained model, unlabeled data set can fit in FSL-BM algorithm using binary input, and  encode all unlabeled data points in same space of trained model as we discussed in section~\ref{PreProcessing}. After using Hashing function, the correct indices is assigned to each data point; and finally, unlabeled has assigned by correct label(s). In feature representation, input binary feature converted to hash keys by meta-feature selected hash function and look at the Supervised hash table. Some data points have more than one label, hence fuzziness logics meaning each data point can be belong to more than one label. As presented in Algorithm~\ref{al3}, main loop is from 0 to $m-1$ where $m$ is the number of test data points and the maximum number of fuzziness is the maximum number of labels each point can be assigned.\\

\begin{figure}[h] \label{fig:workflowedge}
  \centering
    \includegraphics[width=\columnwidth]{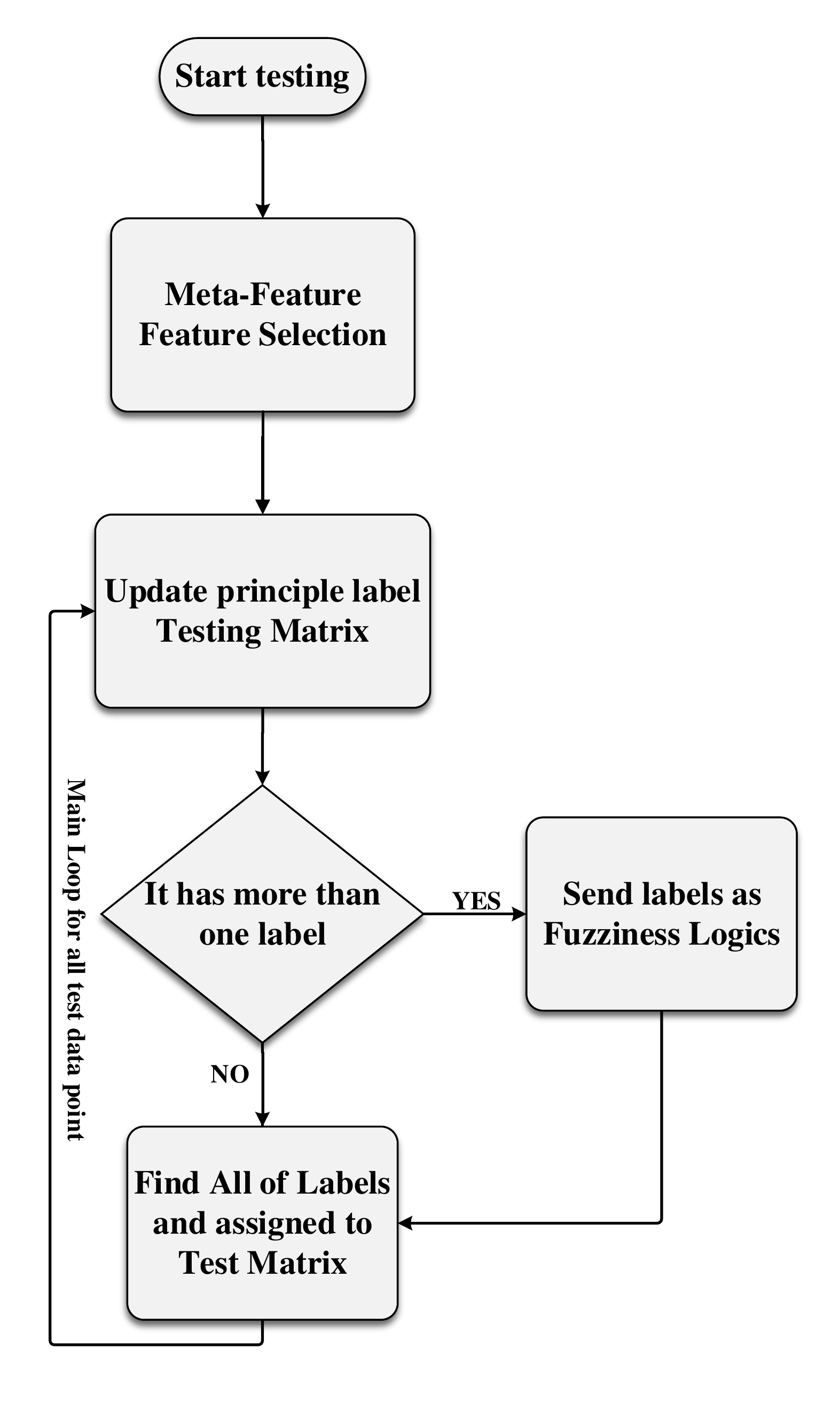}
  \caption{Pipeline of testing our results by supervised training hash table using Hamming Distance}
\end{figure}

\begin{algorithm}[] 
\caption{testing data point by Supervised Learning Hash Tables }\label{al3}
  \begin{algorithmic}[1]
\For{\texttt{$i = 1~ to ~m-1$}}
\For{\texttt{$x = 0$ to max Fuzziness label }}
       
\If{$w_{i,j} \neq null $}\\
\Comment{label of x is available and label is exist in existing training labels list}
\State $Prediction_{i,j} \leftarrow w_{(Hash ~index) ,j}$
\\ \Comment{Add lables to labels list of i}

        \State \textbf{End of if}
        \EndIf
        \State \textbf{End of if}
      \EndFor
      \State \textbf{End of For}
      \EndFor
      \State \textbf{End of For}
  \end{algorithmic}
\end{algorithm}

\section{Experimental Results}\label{Results}
Although time complexity is one of the most important criteria of evaluating the time consumption, the hardware used for implementation and testing the algorithms is pretty essential. Listed time in Table \ref{ta:results} and figure~\ref{random} experimented with single thread implementation, yet using multiple thread implementation would reduce the time. In addition, the other significant factor is memory complexity which is linear in this case, $O(n)$. 

All of the empirical and experimental results of this study shown in Table~\ref{ta:results}  is implemented in a single processor. The source code will be released on GitHub and our lab website that implemented C\textbf{++} and C\# framework. C\textbf{++} and C\#  are utilized for testing the proposed algorithms with a system core $i7$ Central Processing Unit (CPU) with 12 GB memory.

\begin{table}[h]
\centering
\caption{FSL-BM Accuracy}
\label{ta:results}
\begin{tabular}{  c  c  c  c   c  }
\hline
       & \multicolumn{2}{c}{Dataset 1} & \multicolumn{2}{c}{Dataset 2} \\ \hline
       & Accurcy     & Fuzzy Measure    & Accurcy     & Fuzzy Measure    \\ \hline
SVM    & 89.62       & NA               & 90.42       & NA               \\ \hline
FSL-BM & 93.41       & 0.23             & 95.59       & 0.86             \\ \hline
\end{tabular}
\end{table}
\subsection{Data Set}
we test our algorithm in two ways: first empirical data,The data set which is used in this algorithm has 24 binary features. AS regards to table~\ref{ta:results}, we test our algorithm as following data sets: first data-set includes $3,850$ training data points and $2,568$ validation test. And the second data set include $3,950$ data points as training size and $2,560$ validation test. And also we test accuracy and time complexity with random generated data-set as shown in Fig.~\ref{random}.

\subsection{Results}
We test and evaluate our algorithm by two ways which are real dataset from $IMDB$ and \textit{Random Generate Dataset}.~\\
\subsubsection{Results of IMDB dataset}~\\~
Testing a new algorithm with different kinds of datasets is very critical. We test our algorithms and compare our algorithms vs traditional supervised learning methods such as Support Vector Machine (SVM) . The proposed algorithm is validated with two different datasets with 23 binary features. Regarding table \ref{ta:results}, total accuracy of dataset number 1 with 23 binary feature isz;~93.41\%, correct accuracy~:~93.1\%, Fuzziness accuracy: 92.87\%, Fuzziness~:~0.23\% , Boolean Error~:~6.8\%, Fuzzy Error: 7.1\%, and regarding the second data set : Total accuracy is : 95.59\%, correct accuracy~:~94.4\%, Fuzziness accuracy~:~96.87\%, Fuzziness~:~0.86\%, Error:~4.4\%. Regarding table \ref{ta:results} , these results show that Binary Supervised Learning using Hamming Distance has more accurate result in comparison same data set. In first data set, we have~93.41 percent accuracy with FSL-BM while the accuracy in SVM is~89.62 percent and in second data set with 100 training data points, accuracy is~95.59 percent and~90.42 percent.\\~\\
\subsubsection{Results on random online dataset}~\\~
As regards to fig.~\ref{random}, it shows two aspect of FSL-BM 1) this technique has capable for online usage as online learning, and the model works for large volume and big data. the fig.~\ref{random} indicates how this model is learned with big binary meta-feature data sets beside the fast time complexity.

\section{Conclusion and Future Works}
The proposed algorithm (FSL-BM) is effectively suitable for big data stream, where we want to convert our data points to binary feature. This algorithm can be comparable with other same algorithms such as Fuzzy Support Vector Machine (FSVM), and other methods. In this research paper, we presented a novel technique of supervised learning by using Hamming Distance for finding nearest vector and also using meta-feature, meta-knowledge discovery, and meta-learning algorithms is used for improving the accuracy. Hash table and Hash function are used to improve the computational time and the results indicate that our methods have better accuracy, memory consumption, and time complexity. Fuzziness is another factor of this algorithm that could be useful for fuzzy unstructured data-sets which real data-sets could be classified as fuzzy data, once more reiterating each training data point has more than one label.
As a future work, we plan to automate dynamically the number of feature selection process and create meta-feature selection library for public use. This algorithm can be particularly useful for many kinds of binary data points for the purpose of binary big data stream analysis. Binary Features in Fuzzy Supervised Learning is a robust algorithm that can be used for big data mining, machine learning, and any other related field. The authors of this study will have a plan to implement and release the Python, R and Matlab source code of this study, and also optimize the algorithm with different techniques allowing the capability to use it in other fields such as image, video, and text processing.

\bibliographystyle{IEEEtran}
\bibliography{ref}

\end{document}